\documentclass[11pt,a4paper]{article}
\usepackage[hyperref]{eacl2021}
\usepackage{times}
\usepackage{latexsym}

\usepackage{amsmath}
\usepackage{booktabs}

\usepackage{graphicx}

\usepackage{microtype}

\aclfinalcopy 
\def\aclpaperid{1026} 

\title{Coloring the Black Box:\\What Synesthesia Tells Us about Character Embeddings}

\author{Katharina Kann\thanks{~~Equal contribution.}\\
  University of Colorado Boulder\\
  \texttt{katharina.kann@colorado.edu}\\\And
  Mauro M. Monsalve-Mercado\footnotemark[1] \\
  Columbia University \\
  \texttt{~~~mauro.m.monsalve@columbia.edu}\\}

\date{}

\begin{document}
\maketitle
\begin{abstract}
In contrast to their word- or sentence-level counterparts, character embeddings are still poorly understood.
We aim at closing this gap with an in-depth study of English character embeddings.
For this, we use resources from research on grapheme--color synesthesia -- a neuropsychological phenomenon where letters are associated with colors --, which 
give us insight into which characters are similar for synesthetes and how characters 
are organized in color space. 
Comparing 10 different character embeddings, we ask: 
How similar are character embeddings to a synesthete's perception of characters? And how similar are character embeddings extracted from different models? We find that LSTMs 
agree with humans more than transformers. Comparing across tasks, grapheme-to-phoneme conversion results in the most human-like character embeddings. Finally, ELMo embeddings differ from both humans and other models.
\end{abstract}

\section{Introduction}
Neural network models have become crucial tools in natural language processing (NLP) and define the state of the art on a large variety of tasks \cite{wang-etal-2018-glue}. However, 
they are difficult to understand and  are often considered "black boxes".\footnote{This fact has led to the establishment of a workshop with the same name: \url{https://blackboxnlp.github.io}}
This can make their use 
difficult to defend in many settings, for instance in a legal context, and constitutes a barrier for model improvement.
Therefore, 
a lot of research has been dedicated to investigating the information encoded by neural networks. Especially word embeddings, contextualized word representations, and language representation models like BERT \cite{devlin-etal-2019-bert} 
have been exhaustively studied \cite{rogers2020primer}. 

Character embeddings are used for a large set of tasks, 
either as a supplement to word-level input, e.g., for part-of-speech tagging by \citet{plank-etal-2016-multilingual}, or on their own, e.g., for character-level sequence-to-sequence (seq2seq) tasks 
by \citet{kann-schutze-2016-single}. 
Despite this, they have 
not yet been explicitly analysed.
One reason for this might be
that identifying relevant properties to study is more challenging than for their word-level counterparts. However, we argue that, in order to truly shine light into black-box NLP models, it is necessary to understand each and every component of them. 

\begin{figure}[t]
    \centering
    \includegraphics[width=.75\columnwidth]{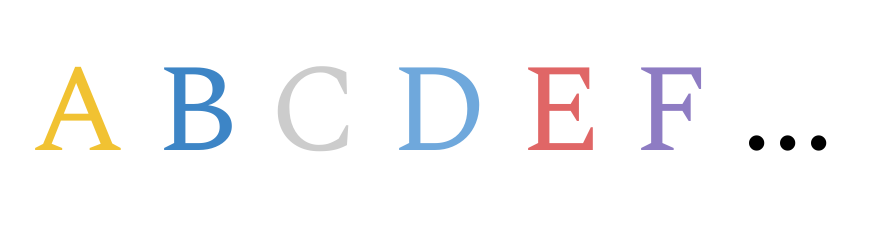}
    \caption{Characters as they might be seen by a grapheme--color synesthete. The colors in this example are randomly chosen.}
    \label{fig:syn}
\end{figure}

In this paper, we perform a detailed study of English character embeddings. Our \textbf{first contribution} is a \textbf{character similarity task} (\S\ref{sec:charsim}) in analogy to the word-based version:
\textit{Do embeddings agree with humans on which pairs of characters are similar?}
While 
e.g., \textit{cat} and \textit{tiger} are generally considered more similar than \textit{cat} and \textit{chair}, this is not trivial for characters. For our annotations, we exploit a phenomenon called \textit{synesthesia}. People with synesthesia, \textit{synesthetes}, share perceptual experiences between two or more senses (\S\ref{sec:syn}). For grapheme--color synesthetes, each letter is associated with a color, which is consistent over a person's life time \cite{eagleman2007standardized}, cf. Figure \ref{fig:syn}. Using a dataset of letters from the English alphabet and the associated colors from 4269 synesthetes \citep{witthoft2015prevalence}, 
we compute differences 
in color space (\S\ref{subsec:color}) 
as a proxy for character similarity.

As our \textbf{second contribution}, we propose
 a set of methods to \textbf{characterize the structure of character embedding matrices} (\S\ref{sec:matrix}). The methods we propose are (i) a clustering analysis, (ii) computing the clustering coefficient, (iii) measuring betweenness centrality, and (iv) computing cut-distances.

Our \textbf{final contribution} is a detailed \textbf{character embedding analysis} (\S\ref{sec:analysis}). We explore 6 types of embeddings, obtained from 4 different architectures trained on 3 different tasks, as well as 4 embedding matrices from pretrained ELMo models \cite{peters-etal-2018-deep}. 
Comparing across models trained on the same task, character embeddings from LSTMs, which,
similar to humans, process input sequentially,
correlate more with human similarity scores than embeddings from transformers. 
Comparing across tasks, character embeddings from language modeling show a surprisingly low correlation with human judgments. In contrast, the correlation is highest for grapheme-to-phoneme  conversion (G2P). This is 
in line with findings that colors perceived by synesthetes are influenced by the sound of each letter \cite{asano2013grapheme}.

\section{Synesthesia}
\label{sec:syn}

Synesthesia is the perceptual phenomenon that two or more sensory or cognitive pathways are co-activated in the brain by stimulating only one of them. For example, for a person with chromesthesia, a common form of synesthesia, hearing a particular sound 
evokes the visual perception of colors \cite{cytowic2018synesthesia,simner2012defining}.

By far the most common form of synesthesia is grapheme-color synesthesia, for which subjects report the perception of colors when seeing numerals or letters  (cf. Figure \ref{fig:syn}). The neural basis behind the phenomenon is still a much debated topic, but evidence suggests it might be the result of increased effective connectivity between the brain areas involved \cite{Ramachandran2001}. For example, visual area V4a in the ventral pathway, often associated with contextual processing and perception of color, has shown to be part of a complex in the fusiform gyrus exhibiting higher cortical thickness, volume, and surface area in synesthetes \cite{HUBBARD2005}. Belonging to this complex and adjacent to the color processing region is the area dedicated to the visual recognition and processing of graphemes, suggesting that a prominence of this type of synesthesia is due to higher region interconnectivity being more probable. 

Additional evidence supports the idea that synesthetic associations often involve the extraction of meaning from a stimulus, suggesting that abstract constructions play a major role in the associations formed \cite{Mroczko-Wasowicz2014}. For cases where language is involved, semantics may in part underlie color representations of words and graphemes in V4. 

The study of synesthesia presents a unique opportunity to understand the neural basis of cognitive models of language \cite{Ramachandran2001.2,Simner2007-SIMBPS}. Here, grapheme--color synesthesia serves as a window to look at how the brain represents individual characters.

\section{A Character Similarity Task}
\label{sec:charsim}
\subsection{Motivation: Word Similarity} 
Our first contribution is a character-level analogue of the word similarity task, 
an intrinsic evaluation method for word embeddings. It 
consists of judging the similarity of pairs of words, e.g., of \textit{cat} and \textit{tiger}. 
To obtain a gold standard for this task, human annotators 
assign similarity scores 
to a list of word pairs. This is not always trivial: are \textit{cat} and \textit{bird} more or less similar than \textit{cat} and \textit{fish}? However, people tend to agree on general tendencies. 

Word embeddings are evaluated on similarity datasets as follows: The similarity -- usually cosine similarity -- of all pairs of words 
is computed. 
The agreement between models and human annotations is then quantified as the correlation between the two vectors of scores.
Word embeddings with a higher performance on word similarity tasks are expected to perform better on downstream tasks, since they encode valuable information about words and the relationships between them.

\subsection{From Word to Character Similarity} 
In order to design a \textit{character} similarity task, we require a gold standard. 
However, we are not likely to get a meaningful answer when asking people if \textit{B} and \textit{J} are more similar than \textit{C} and \textit{Q}. 

We solve this problem by looking at how different characters are represented by grapheme--color synesthetes in color space. This tells us how similar characters are perceived to be without having to ask explicitly.
We leverage a dataset collected by \newcite{witthoft2015prevalence}, which consists of letter-to-color mappings collected from 4269 synesthetes and 
compute pair-wise perceptually uniform distances between characters. Analogously to the word-level version of the task, we then take as our gold standard the average over all annotators. For embeddings, we compute cosine similarities between character vectors.

We differ from the word similarity task in an important detail: we do not evaluate the \textit{quality} of embeddings. Instead, we aim to understand character embeddings 
by assessing how \textit{similar to the human perception of characters} they are. In particular, we do not necessarily expect embeddings which score higher on our task to perform better on downstream applications.

\subsection{Human Perception of Color Differences}
\label{subsec:color}
To motivate the distance metric we use, we  
first summarize human perception of color.
The visual process of color perception starts in the retina, where three types of light-sensitive receptors 
are tuned to respond broadly around three distinct frequencies in the visible spectrum. However, color perception is not simply reduced to combinations of these physiological responses. The human brain can perceive the same physical light frequency as a different color depending on a plethora of contextual markers, mostly due to further processing upstream of the retina into high-level visual cortical areas, where associations are formed between environmental cues and the object being visualized.

Important contextual properties of color affecting perception, often discussed in color theory as the Munsell 
(or HSL) color system, 
are its hue, 
saturation, 
and lightness. Combinations of these properties are interpreted in a \textit{highly non-linear} manner by cortical areas tasked with color perception. Not surprisingly, simple metrics for color comparison, such as the Euclidean distance in RGB space, perform poorly in situations involving a color discrimination task. Constructing perceptually uniform metrics that deal with these perceived non-linearities is an active field of research. One standard metric for perceptual color comparison is the CIEDE2000 color difference \cite{Sharma2005CIEDE2000}, which includes several correction factors for a modified Euclidean metric on HSL space.
We employ CIEDE2000 in our analysis.

\section{Analysis of the Distance Matrix}
\label{sec:matrix}
CIEDE2000 allows us to obtain pair-wise distances between all letters in the alphabet by computing color differences for the associated colors as perceived by synesthetes. 
The character similarity task then compares \textit{vectors} of pair-wise distances. However, we can gain additional insight 
from the distance \textit{matrices}, which represent fully connected networks whose nodes are the letters in the alphabet and whose edges are weighted by the pair-wise differences. Thus, we further 
propose four well-established methods from network theory \cite{newman2018networks} for the analysis of character embeddings and human difference matrices.

\textbf{Clustering analysis.}
To characterize the global structure of the network, we first propose Ward's variance minimization algorithm to identify 
clusters, i.e.,
groups of letters that are similar to each other, but far away from other letters. Ward's algorithm is part of a family of hierarchical 
clustering algorithms whose objective function aims at minimizing the variance within clusters \cite{Ward1963}. Starting from a forest of clusters (initially single nodes), the algorithm evaluates the Ward distance $d$ between a 
new cluster $u$ made up of clusters $s$ and $t$, and a third cluster $v$ not used yet, as
\begin{equation}
    d(u,v)=\sqrt{a + b - c},
\end{equation}
with 
\begin{align}
    a &= \frac{|v|+|s|}{T}d(v,s)^2, \\
    b &= \frac{|v|+|t|}{T}d(v,t)^2, \\
    c &= \frac{|v|}{T}d(s,t)^2,
\end{align}
where $T:=|s|+|t|+|v|$, and $|\cdot|$ is the size of the cluster. If $u$ is a good cluster, then $s$ and $t$ are removed from the forest and the algorithm continues until only one cluster is left. Finally, the \textit{number of clusters}, \textit{their sizes}, and the \textit{Ward distances between clusters}  characterize the global structure of the network.

\textbf{Clustering coefficient.}
The clustering coefficient provides a
way to measure the degree to which nodes in a network cluster together. For a binary network, the local version represents the fraction of the number of pairs of neighbors of a node that are connected, over the total number of pairs of neighbors of said node, and it measures the influence of a node on its immediate neighbors. Several generalizations for weighted networks have been proposed \cite{saramaki2007}. Here, we use the average of the weights for neighbors in the subgraph of a node $u$
\begin{equation*}
    c_u = \frac{1}{deg(u)(deg(u)-1))}
      \sum_{vw} (\hat{w}_{uv} \hat{w}_{uw} \hat{w}_{vw})^{1/3},
\end{equation*}
where $\hat{w}_i$ denotes weight $w_i$ normalized by the maximum weight in the network, and $deg$ is the degree of the node (the sum of its edges' weights). The average over all nodes is used as a proxy for the overall level of clustering within the network.

\textbf{Betweenness centrality.}
Different concepts of centrality attempt to capture the relative importance of particular nodes in the network. One such concept, \textit{betweenness}, measures the extent to which a node lies on the shortest path between pairs of nodes \cite{BRANDES2008136}. In a sense, it generalizes the clustering coefficient from a measure of the local influence of a node to immediate neighbors to the whole network. In particular, it accounts for nodes that connect two different clusters while not being a part of either. Betweenness centrality is computed as the fraction of all-pairs shortest paths that pass through a particular node $u$
\begin{equation*}
    B(u) =\sum_{s,t} \frac{n(s, t|u)}{n(s, t)},
\end{equation*}
where the sum is over all nodes $s$ and $t$ in the network, and $n$ counts the number of shortest paths between two nodes, optionally taking into account if it passes through $u$.

\paragraph*{Cut-distance.}
The fourth and last approach we propose to characterize our matrices of character distances is to employ a matrix norm called cut-norm.
It is widely used in graph and network theory and has been shown to capture global features such as clustering and sparseness \cite{Frieze1999}. The cut-norm of a matrix $A=(a_{ij})_{i\in M, j\in N}$ is defined as
\begin{equation*}
    ||A||_c := \textrm{max}\left\{ \frac{\left| \sum_{i \in I} \sum_{j \in J} a_{ij}\right|}{|I||J|} : 
    {I\subset M \atop J\subset N}\right\},
\end{equation*}
i.e., the maximum over all possibles sub-matrix arrangements is taken as the norm. In practice, we compute it using an efficient implementation\footnote{\url{https://pypi.org/project/cutnorm/}} that relies on Grothendieck's inequality for an approximation \cite{Alon2004Cut-Norm,wen-etal-2013-extracting}. In addition, the norm naturally gives rise to a distance metric $d_c(A,B):=||A-B||_c$ that allows us to compare pairs of distance matrices directly.

\section{Models and Tasks}
\label{m_and_t}
We now describe the tasks and model architectures we employ to train different character embeddings.

\subsection{Tasks}
\textbf{Language modeling. } The task of language modeling consists of computing a probability distribution over all elements in a predefined vocabulary, given a sequence of past elements. Language models can either be used to assign a probability to an input sequence or to generate text by sampling from the probability distributions. 
We train language models on the character level, i.e., the vocabulary consists of the English alphabet.

All our language models are trained on wikitext-103.\footnote{\url{https://s3.amazonaws.com/research.metamind.io/wikitext/wikitext-103-v1.zip}} We use the provided training, development, and test splits. The training set consists of roughly 1 million tokens.

\textbf{Morphological inflection. } 
In languages that exhibit rich inflectional morphology, words inflect: grammatical information like number, case, and tense are incorporated into the word itself. For instance, \textit{wrote} is the inflected form of the English lemma \textit{write}, expressing past tense.

The task of morphological inflection consists of mapping a lemma to an inflected form which is defined by a set of morphological tags.
Morphological inflection is typically being cast as a character-level seq2seq task, where the characters of the lemma together with the morphological tags are the input, and the characters of the inflection are the output \cite{kann-schutze-2016-single}:
\begin{align}
\texttt{PST w a l k} ~ \rightarrow ~ 
\texttt{w a l k e d} \nonumber
\end{align}

\noindent We train our inflection models on the $10,000$ English training examples provided by \citet{cotterell-etal-2017-conll} and use the corresponding development and test sets with $1,000$ examples each. 

\textbf{Grapheme-to-phoneme conversion. } 
Given a word's spelling, G2P consists of generating an (IPA-like) representation of its pronunciation:
\begin{align}
\texttt{p r e t t i e r} ~ \rightarrow ~ 
\texttt{P R IH T IY ER} \nonumber
\end{align}
It has been shown that similar-sounding letters tend to be associated with similar synesthetic colors  \cite{asano2013grapheme}. Hence, we assume that
the embedding space induced by this task could be similar to human perception of characters.

We train all G2P models on examples extracted from the CMU Pronouncing Dictionary.\footnote{http://www.speech.cs.cmu.edu/cgi-bin/cmudict} Our training, development, and test sets consist of 114,399, 5447, and 12,855 examples, respectively.\footnote{We use the splits provided at \url{https://github.com/microsoft/CNTK/tree/master/Examples/SequenceToSequence/CMUDict/Data}.}

\subsection{Architectures}
To isolate the effects of task and model architecture, we train different architectures for each task.
All test set performances are shown in Table \ref{tab:sanity_check}. 
We train 50 models with different random seeds for seq2seq tasks,
and 10 instances for language models. For our analysis, we look at the input embeddings and average pair-wise distances over models for each group.  
All models have been trained on an NVidia Tesla K80 GPU.

\textbf{LSTM seq2seq architecture. } Our first architecture is a seq2seq model similar to that by \newcite{bahdanau2015neural}. 
It consists of a bi-directional long short-term memory \citep[LSTM; ][]{hochreiter1997long} encoder and an LSTM decoder, which are connected via an attention mechanism. 
We apply it on the character level.

We train this architecture on morphological inflection (\textbf{$\textrm{Infl}_\textrm{LSTM}$}) and G2P (\textbf{$\textrm{G2P}_\textrm{LSTM}$}), using the fairseq sequence modeling toolkit for our implementation.\footnote{\url{https://github.com/pytorch/fairseq}} 
All embeddings and hidden states are 100-dimensional, and both encoder and decoder have 1 hidden layer. For training, we use an Adam optimizer \cite{kingma2014adam} with an initial learning rate of 0.001, dropout with a coefficient of $0.3$, and a batch size of 20. To account for different training set sizes, we train our model for G2P for 15 and our model for inflection for 100 epochs.

\textbf{Transformer seq2seq architecture. } 
We further experiment with a transformer seq2seq architecture \cite{vaswani2017attention}. Similar to the LSTM seq2seq model, this  architecture consists of an encoder and a decoder which are connected by an attention mechanism. However, the encoder and the decoder 
consist of combinations of feed-forward and attention layers instead of LSTMs. 

We apply this architecture to  morphological inflection (\textbf{$\textrm{Infl}_\textrm{T}$}) and G2P (\textbf{$\textrm{G2P}_\textrm{T}$}), and implement the models using the fairseq toolkit.
All embeddings have 256 dimensions, and hidden layers are 1024-dimensional. Both encoder and decoder have 4 layers, and use 4 attention heads.
We employ an Adam optimizer with an initial learning rate of 0.001 for training, together with dropout with a coefficient of $0.3$, and a batch size of 400. We train our models for G2P for 30 epochs and our models for morphological inflection for 100 epochs.

\textbf{LSTM language model architecture. }
We also experiment with an LSTM language model (\textbf{$\textrm{LM}_\textrm{LSTM}$}). This architecture consists of a unidirectional LSTM, and it receives the last generated character as input at each time step.

Our implementation is based on the official pytorch LSTM language model example.\footnote{\url{https://github.com/pytorch/examples/tree/master/word_language_model}} 
We use the default hyperparameters except for the following: our embeddings and LSTM hidden states are 512-dimensional, we use 2 hidden layers, and we train for 2 epochs 
with a batch size of 64. 

\textbf{Transformer language model architecture. }
Our last architecture is a transformer language model (\textbf{$\textrm{LM}_\textrm{T}$}). 
Like the LSTM language model, it receives previously generated characters as input and computes a probability distribution over the character vocabulary.

Again, we use the fairseq toolkit for our implementation, and employ the default hyperparameters for the transformer language model.
We use an Adam optimizer with an initial learning rate of 0.0005 for training, and dropout with a coefficient of 0.1. This model is trained for 3 epochs.

\begin{table}[t] 
  \setlength{\tabcolsep}{3.8pt}
  \small
  \centering
  \begin{tabular}{ c c c c c c}
    \toprule
    \textbf{$\textrm{Infl}_\textrm{LSTM}$} &
    \textbf{$\textrm{Infl}_\textrm{T}$} & \textbf{$\textrm{G2P}_\textrm{LSTM}$} &
    \textbf{$\textrm{G2P}_\textrm{T}$} &
    \textbf{$\textrm{LM}_\textrm{LSTM}$} &
     \textbf{$\textrm{LM}_\textrm{T}$} \\
    \midrule
    0.95 & 0.94 & 0.64 & 0.62 & 3.37& 2.62 \\
    50 & 50 & 50 & 50 & 10 & 10 \\
    \bottomrule
  \end{tabular}
  \caption{\label{tab:sanity_check} \textbf{Top}: accuracy for inflection and G2P and 
  character-level perplexity for language modeling. 
  \textbf{Bottom}: number of model instances. Results are averaged over all runs; models are described in the text.
  } 
\end{table}

\subsection{Pretrained Models}
We further analyze the character embeddings of ELMo models \cite{peters-etal-2018-deep}. 
ELMo models are pretrained networks, aimed at producing contextualized word embeddings for use in downstream NLP tasks. 
The model architecture consists of a convolutional layer over character embeddings, whose output is then fed into a 2-layer bidirectional LSTM. 
ELMo models are trained with a bidirectional language modeling objective.

We experiment with 4 English models which are available online:\footnote{\url{https://allennlp.org/elmo}} small (\textbf{$\textrm{ELMo}_\textrm{s}$}), medium (\textbf{$\textrm{ELMo}_\textrm{m}$}), original (\textbf{$\textrm{ELMo}_\textrm{o}$}) and original-5.5B (\textbf{$\textrm{ELMo}_\textrm{l}$}). Those models differ in the number of their parameters and the amount of text they have been trained on: $\textrm{ELMo}_\textrm{s}$ and $\textrm{ELMo}_\textrm{m}$ have 13.6 million and 28 million parameters, respectively. Both $\textrm{ELMo}_\textrm{o}$ and $\textrm{ELMo}_\textrm{l}$ have 93.6 million parameters. All models except for $\textrm{ELMo}_\textrm{l}$ have been trained on the 1 Billion Word Benchmark.\footnote{\url{http://www.statmt.org/lm-benchmark/}} $\textrm{ELMo}_\textrm{l}$ has been trained on a combination of Wikipedia and news crawl data, which together result in a dataset of 5.5 billion tokens.

\section{Character Embedding Analysis}
\label{sec:analysis}

\subsection{Results on the Character Similarity Task}
\label{subsec:sim}

\begin{figure}
    \centering
    \includegraphics[width=\columnwidth]{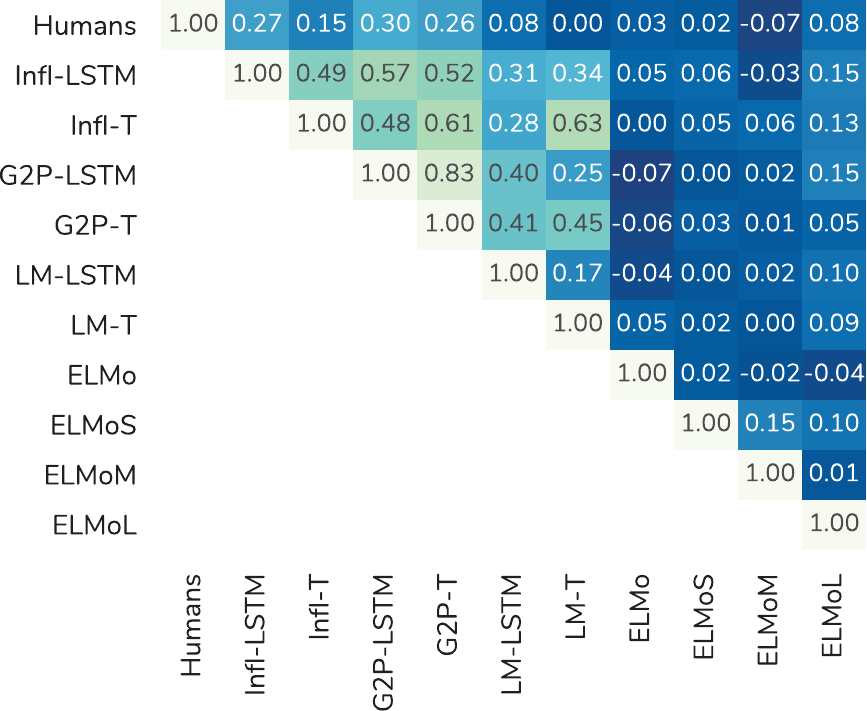}
    \caption{Pearson correlation between the vector of averaged human character differences and distance vectors according to character embeddings.}
    \label{Correlations}
\end{figure}

The Pearson correlation of all models with human judgements as well as 
with each other is shown in Figure \ref{Correlations}. $\textrm{G2P}_\textrm{LSTM}$ shows  the highest correlation with 0.30, while $\textrm{LM}_\textrm{T}$ is not correlated at all, and $\textrm{ELMo}_\textrm{m}$ obtains the strongest negative correlation with  $-0.07$.

More generally, we see that most models are correlated with each other (between $0.17$ and $0.83$), with the exception of the ELMos: the predictions of $\textrm{ELMo}_\textrm{l}$ are the only ones which have a Pearson correlation $\geq0.1$ with those of some other models. 

Comparing to human scores, we find the following patterns: embeddings from seq2seq tasks show a higher correlation than embeddings from language models. Even ELMo models, which are trained on large amounts of text, obtain a maximum correlation of $0.08$. Thus, we conclude that language modeling does not result in embeddings which perform well on our character similarity task. Embeddings for G2P correlate more strongly with human character perception than embeddings from inflection, when comparing identical architectures. This is noteworthy, since colors perceived by synesthetes are supposed to be influenced by the sound of each letter \cite{asano2013grapheme}, similar to the embeddings for G2P. 
Finally, comparing across architectures, we see that LSTM models correlate stronger with human judgments than transformer models, which is in line with the common understanding that recurrent neural networks might be better models of human cognition. 

\begin{figure}
    \centering
    \includegraphics[width=\columnwidth]{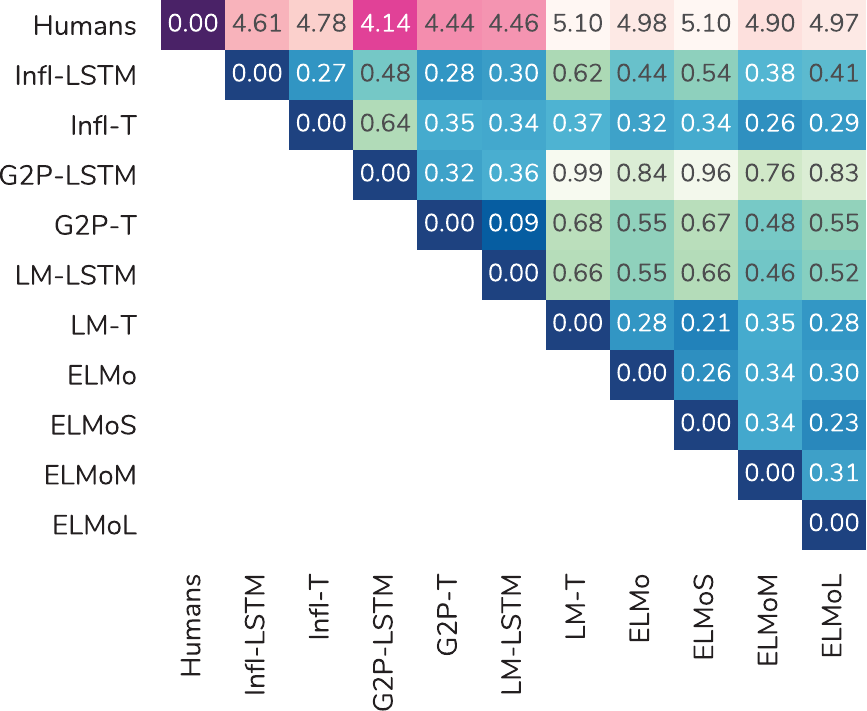}
    \caption{Cut-distance between the average human distance matrix and all character embeddings.}
    \label{CutDistances}
\end{figure}
\subsection{Comparing Humans and Machines}

\begin{figure*}
    \centering
    \includegraphics[width=.95\textwidth]{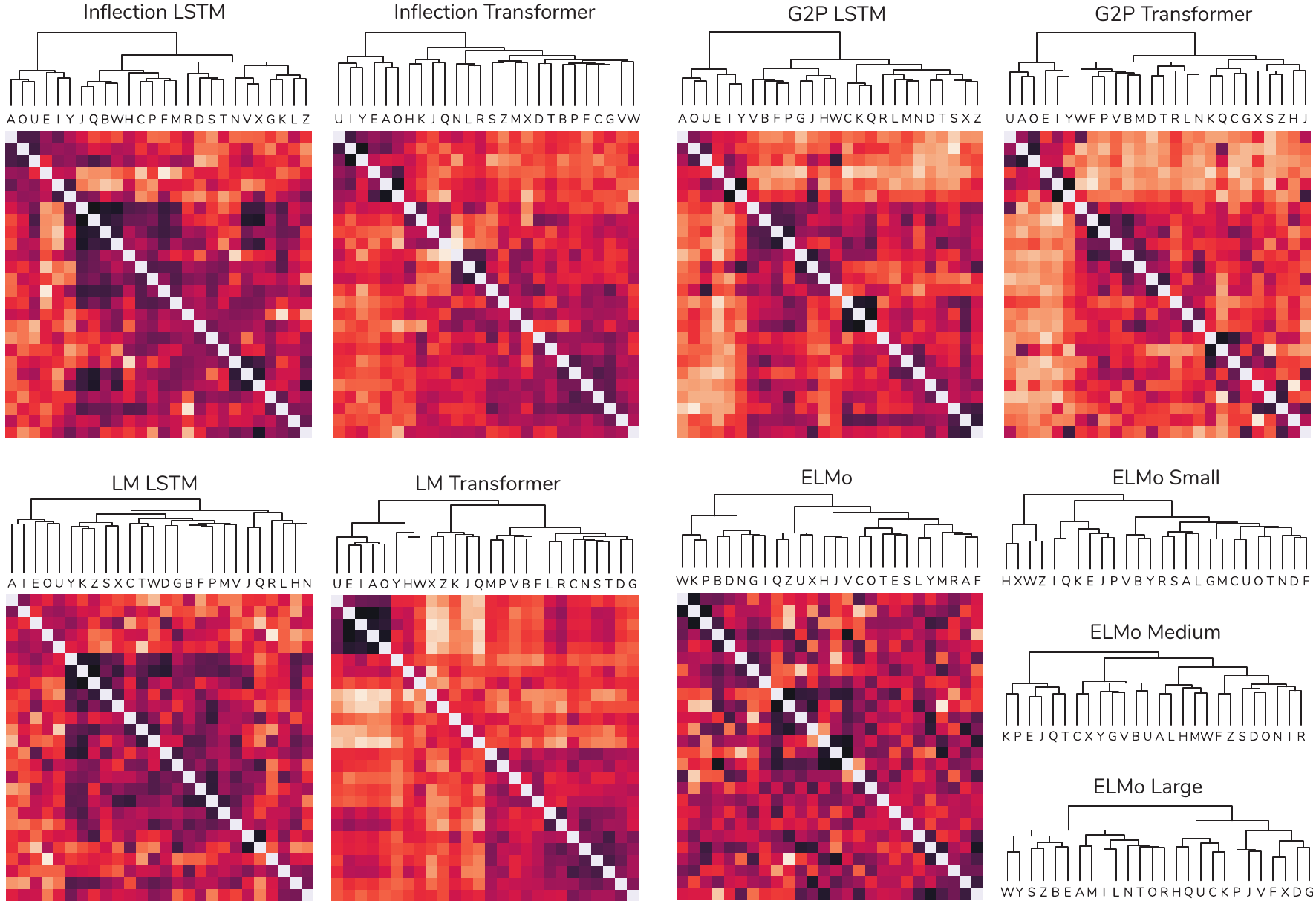}
    \caption{Distance matrices and corresponding dendrograms reveal the cluster structure of character embeddings. Darker colors depict small distances (high similarity) between pairs. 
    Dendrograms summarize the cluster structure, with the height of horizontal lines depicting the Ward distance between the corresponding clusters being joined.}
    \label{BigAssFig}
\end{figure*}

\textbf{Clustering analysis. } First, we look at the global structure of all character embeddings 
(cf. Figure \ref{BigAssFig}). All but the ELMo models exhibit a marked separation between a tight cluster of vowels (\texttt{AEIOU}) or extended vowels (+\texttt{Y}), which are highly similar, and the rest of the alphabet. 
In contrast, this distinction is not found for humans, neither for individuals nor the average distance matrix.
Despite this, the human average does present a clear global structure (cf. appendix for details). One clear cluster is \texttt{BDGJKMNPQRVWXZ}, with the particularly close pairs \texttt{MN} and \texttt{XZ}, perhaps due to the letters' shape, sound, or  
proximity in the alphabet. This cluster is far away from the letters in \texttt{AES}, which themselves do not form a cluster. Another cluster, on average far form the first, is formed by the letters \texttt{CILOUY}.

\begin{figure}[t]
    \centering
    \includegraphics[width=\columnwidth]{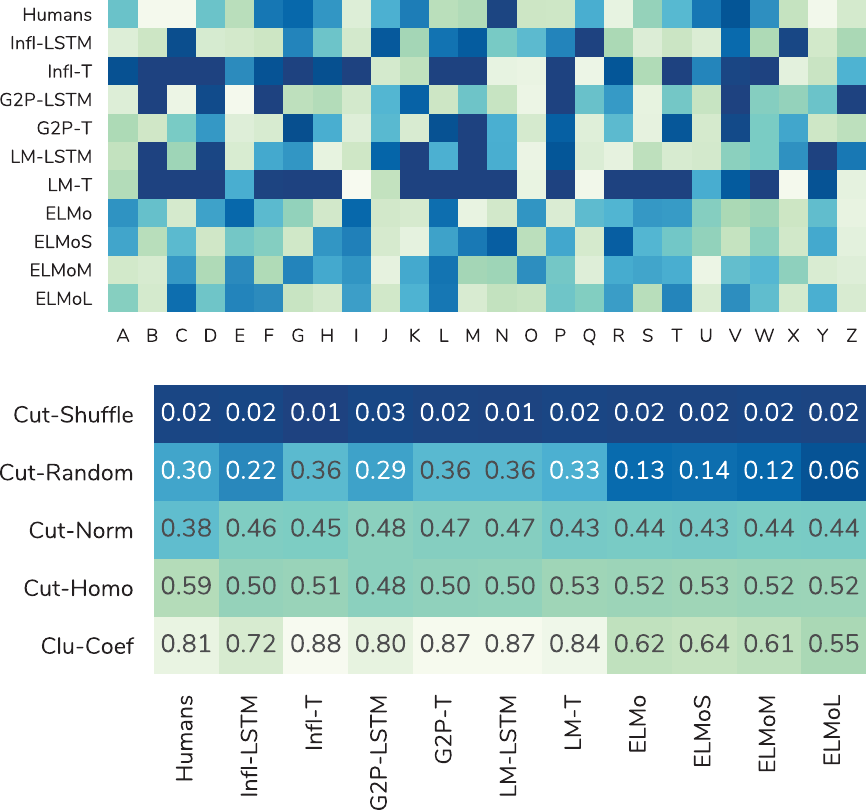}
    \caption{Comparison measures for models of character embedding. {\bf Top}: Betweenness centrality.  {\bf Bottom}: Cut-distances and cluster coefficients for the average human distance matrix and all character embeddings.}
    \label{CutMachines}
\end{figure}
Apart from the clear separation between vowels and consonants, character embeddings exhibit rich additional structure (cf. Figure \ref{BigAssFig}).
For $\textrm{Infl}_\textrm{LSTM}$, the cluster \texttt{BCFHJMPQW} contains the tighter sub-cluster \texttt{BJQW} that is similar to the letters \texttt{GKVX}.
In contrast, $\textrm{Infl}_\textrm{T}$ has the two small clusters \texttt{HJKQ} and \texttt{LNR} far from each other, and a less clear-cut structure among the rest of the consonants. 

Embeddings from $\textrm{G2P}_\textrm{LSTM}$ exhibit a structure 
clearly connected to the G2P task.
The distance network has many small clusters of similar sounding letters, e.g.,
\texttt{UW,IJ,JY,GJ,CKQ,SXZ}, and \texttt{BFPV}.
$\textrm{G2P}_\textrm{LSTM}$'s counterpart, $\textrm{G2P}_\textrm{T}$, produces similar groupings, but displays a more homogeneous structure overall. 

$\textrm{LM}_\textrm{LSTM}$ has the tight cluster \texttt{KSXYZ} within the larger \texttt{BCDFGKMPSTVWXYZ} cluster, and weaker connections between other letters.
$\textrm{LM}_\textrm{T}$ has a clearer cluster structure, showing the strongest separation between vowels and consonants, especially from the non-cluster \texttt{JKQXZ}, which is also well-separated from other consonants.
ELMo models are outliers in that they do not present a clear global structure: only loose clusters can be identified. 

\textbf{Cluster coefficient. } Looking at the cluster coefficients (cf. Figure \ref{CutMachines}, last row) we also see a difference between ELMo embeddings and the other models: all models except for the ELMos have cluster coefficients between 0.72 and 0.88 and, thus, close to the human average of 0.81. In contrast, the cluster coefficients for ELMo embeddings are between $0.64$ for $\textrm{ELMo}_\textrm{s}$ and $0.55$ for $\textrm{ELMo}_\textrm{l}$.

\textbf{Betweenness centrality.} The local values of betweenness centrality (cf. Figure \ref{CutMachines})
show 
the rich structure of the similarity matrices for most embeddings. For all but the ELMo models, the majority of letters have either extremely high or low levels of betweenness. In particular vowels tend to occupy prominent places in the network structure. Humans, in contrast, are more similar to ELMo embeddings.

\textbf{Cut-distance.} Looking at cut-distances (cf. Figure \ref{CutMachines}), we find that the structure of ELMo  embeddings is significantly more similar to a random matrix than that of the other embeddings. 
The cut-distances (cf. Figure \ref{CutDistances}) between humans and embeddings largely agree with the conclusions from Section \ref{subsec:sim} -- $\textrm{G2P}_\textrm{LSTM}$ and the ELMo models are respectively the most similar and dissimilar --, even though correlation for node-to-node similarities does not necessarily imply a similar global structure.

\section{Related Work}
\textbf{Neural network analysis. }
A lot of ink has been spilled on what neural network models learn and how. For instance, \citet{zhang-bowman-2018-language} investigated different pretraining objectives on their ability to induce syntactic and part-of-speech information.  \citet{pruksachatkun2020intermediate} studied model performance on probing tasks to investigate what models learn from intermediate-task training. \citet{belinkov-etal-2017-neural} explored what neural machine translation models learn about morphology. 

Other work created test sets to evaluate specific linguistic model abilities. \citet{linzen-etal-2016-assessing} made 
a dataset
to investigate the ability of neural networks to detect mismatches in subject–verb agreement in the presence of distractor nouns. \citet{warstadt2019blimp} created a benchmark called BLiMP to assess the ability of language models to handle specific syntactic phenomena in English. \citet{mueller2020cross} introduced a similar suite of test sets in English, French, German, Hebrew and Russian, also focusing on syntactic phenomena. Similarly, \citet{xiang2021} presented CLiMP, a benchmark for the evaluation of Chinese language models.

Besides that, attention mechanisms \cite{bahdanau2015neural} in neural models have been common subjects of investigation. \citet{jain-wallace-2019-attention} claimed that "attention is not explanation", to be later on challenged by \citet{wiegreffe-pinter-2019-attention}, who argued that "attention is not not explanation". However, the relationship  between inputs, attention weights, and outputs is still poorly understood.

Furthermore, our work is  related to research on which information is learned and how information is encoded by so-called language representation models, e.g., BERT \cite{devlin-etal-2019-bert} or RoBERTa \cite{liu2019roberta}. Similar to attention in other models, attention in BERT has been investigated exhaustively.  \citet{clark2019does} found that it captures substantial syntactic information, and \citet{vig2019visualizing} built a visualization tool for the attention mechanism. 
\citet{hewitt-manning-2019-structural} evaluated whether syntax trees could be recovered from BERT or ELMo's word representation space. An overview of over 40 different studies of BERT can be found in \citet{rogers2020primer}.

\textbf{Embedding analysis. }
The research which is closest 
to our work investigates which information is captured by different types of embeddings, often by training a classifier to predict certain features of interest. 
For instance, 
\citet{kann-etal-2019-verb} used a classifier-based approach to examine whether word and sentence embeddings encode information about the frame-selectional properties of verbs. \citet{ettinger-etal-2016-probing} investigated the grammatical information contained in sentence embeddings regarding multiple linguistic phenomena.
\citet{qian-etal-2016-investigating}
mapped a dense embedding to a sparse linguistic property space to explore the contained information. \citet{bjerva-augenstein-2018-phonology} studied language embeddings.

Different word similarity datasets have been used for word embedding evaluation, for instance RG-65 \cite{10.1145/365628.365657}, WordSim-353 \cite{10.1145/503104.503110}, or SimLex-999 \cite{hill-etal-2015-simlex}.

In contrast to the work in this paragraph, which was concerned with word or sentence embeddings, we aim at understanding \textit{character} embeddings.

\section{Conclusion}
In this paper, we performed an in-depth analysis of character embeddings extracted from various character-level models for NLP.
We leveraged resources from research on grapheme--color synesthesia -- a neuropsychological phenomenon where letters are associated with colors --, to construct a dataset for a character similarity task. We further performed an analysis of networks representing characters as nodes and similarities as edge weights 
to understand how characters are organized by human synesthetes in comparison to character embeddings. Analysing 10 different character embeddings, we found that LSTMs agreed with humans more than transformer models. Comparing different tasks, G2P resulted in embeddings more similar to human character representions than inflection and, by a wide margin, language modeling. ELMo embeddings differed from humans and other models in that they exhibited no clear structure.

\section*{Acknowledgments}
We would like to thank Manuel Mager and the anonymous reviewers for their feedback on this work. Mauro M. Monsalve-Mercado acknowledges financial support from the Center for Theoretical Neuroscience at Columbia University through the NSF NeuroNex Award DBI-1707398 and The Gatsby Charitable Foundation.

\bibliography{anthology,eacl2021}
\bibliographystyle{acl_natbib}

\appendix

\onecolumn

\section*{Appendix A: Statistics}
\label{sec:appB}
 \begin{table}[h]
  \setlength{\tabcolsep}{2pt}
  \small
  \centering
  \begin{tabular}{l}
    \toprule
    X--Z, M--N, R--X, K--X, H--N, M--R, O--U, M--W, V--Z, O--Q \\
    \midrule
    B--Y, B--C, C--R, C--M, M--Y, R--Y, C--X, I--R, P--Y, C--Z \\
    \bottomrule
  \end{tabular}
  \caption{The most similar character pairs in descending order (top) and the most dissimilar character pairs in ascending order (bottom) for human synesthetes.}
  \label{tab:stats}
\end{table}

\section*{Appendix B: Analysis of Color Differences of Characters as Perceived by Synesthetes}
In additional to the most important findings mentioned in the main part of this paper, we further present an in-depth study of the character similarities according to human synesthetes in this section.

\begin{figure*}[t]
    \centering
    \includegraphics[width=\textwidth]{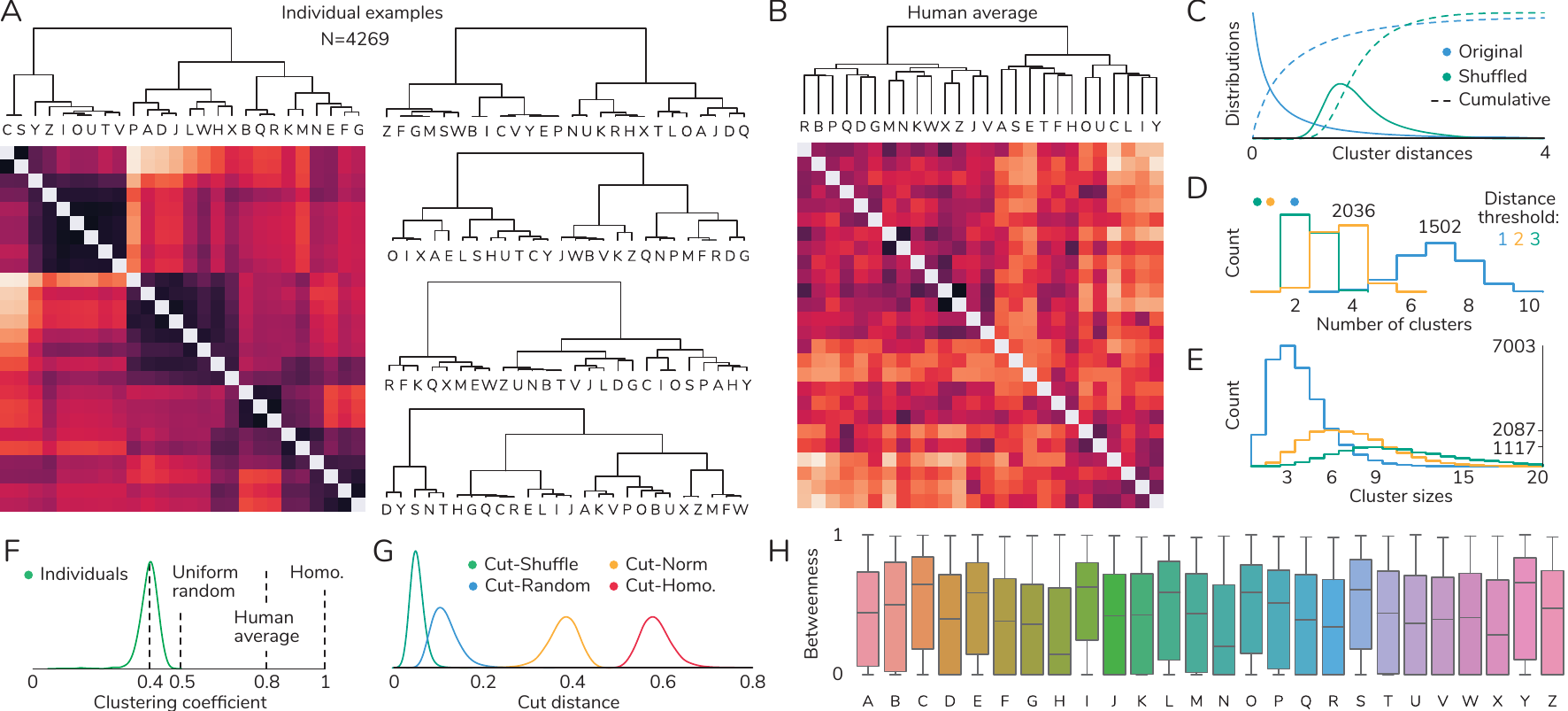}
    \caption{Network analysis of human synesthetes. {\bf (A)} Five randomly chosen examples illustrate the typical distance matrices for individual synesthetes (N=4269). The dendrogram summarizing the clustering structure is presented on top of its corresponding distance matrix for one example (left). The dendrogram is a tree-level representation of the identified clusters, where the height of each horizontal line represents the Ward distance between the selected pair of clusters. 
    {\bf (B)} The clustered distance matrix and corresponding dendrogram of the node-to-node average for all human synesthetes. {\bf (C)} Distribution of all the distances between clusters for each distance matrix pooled over all individuals. Cluster distances correspond to the heights of all horizontal lines in each dendrogram. For comparison, the distribution of cluster distances corresponding to all shuffled distance matrices is presented, together with their cumulative distributions. {\bf (D)} Histograms of the number of clusters in each distance matrix pooled over all individuals for three different cutoff cluster distances. {\bf (E)} Histograms of the size of the clusters found for the procedure in (D). {\bf (F)} Distribution of the average clustering coefficient for each distance matrix pooled over all individuals. For comparison the average clustering coefficient for a random distance matrix, for the human average of (B), and for an homogeneous distance matrix are also marked by dashed lines. {\bf (G)} Distribution of cut-distances between all the human distance matrices and the zero matrix (cut-norm), a random distance matrix (average over 100 iterations), a homogeneous distance matrix, and shuffle versions of themselves (average over 100 iterations). {\bf (H)} Distribution of betweenness centrality for each letter pooled over all individual synesthetes.}
    \label{Humans}
\end{figure*}

\textbf{Clustering Analysis. }
For each individual, we compute a  character difference matrix using CIEDI2000 (normalized to values between 0 and 1). 
We then use Ward's hierarchical clustering algorithm to explore hidden structural features.
Several examples suggest that individual synesthetes tend to represent certain groups of letters with closely matching colors, since their perceptual color differences tend to form tight clusters (Figure \ref{Humans}A), as opposed to the node-to-node average over the entire population of synesthetes (Figure \ref{Humans}B). To make this finding concrete, we compute several measures aimed to address the degree of clustering of each network. 

First, we compute the distances between identified clusters in the distance matrix for each individual (Figure \ref{Humans}C). Pooled over the whole population, the distribution of cluster distances reveal an over-representation of small distances when compared to shuffled data. Over-represented small clustered distances and large jumps to big distances implies the existence of just a handful of tight clusters with small within-cluster distances and larger inter-cluster distances, as suggested by the dendrogram examples in Figure \ref{Humans}A. 

Imposing a cut-off cluster distance for the dendrograms, we effectively select the cluster structure of the distance matrix. Although several methods for selecting the cut-off have been studied, they often strongly depend on the nature of the dataset in question. Using three largely different cut-offs we find robustly that each distance matrix encodes only a handful of tight clusters (cf. Figures \ref{Humans}D, E), typically around 3 to 5 clusters with an average size of 6 letters per cluster.

\textbf{Clustering coefficient. } Next, we 
compute the local clustering coefficient for each node in each distance matrix and observe no strong differences among nodes belonging to the same matrix. For each matrix, we average the cluster coefficient over all nodes and look at their distribution by pooling over all individuals (cf. Figure \ref{Humans}F). This reveals a narrow distribution of low clustering coefficients peaked around $0.4$, implying a small average distance of any node to its neighbours. For comparison, we compute the clustering coefficient for a random uniform distance matrix (symmetric with zeroes in its diagonal) whose coefficient places at $0.5$ (averaged over 100 iterations), and also for a homogeneous distance matrix (all entries are equal with a zero diagonal) with a coefficient of $1$. Moreover, we repeat the analysis for the node-to-node average distance matrix and show a relatively higher coefficient ($0.8$), implying higher distances on average for any node with respect to its neighbors.

\textbf{Betweenness centrality. } Next, we examine the betweenness centrality of nodes (cf. Figure \ref{Humans}H). We compute this measure on the auxiliary similarity matrix $0.5-d$, where $d$ is a distance matrix, in order to interpret high betweenness values as the most important nodes. The distribution over all individuals reveals the top-ranking nodes as \texttt{CEILOSY}. These are the nodes found to be in most shortest paths between pair of nodes, and possibly lie at the intersection between otherwise separate clusters.

\textbf{Cut-distance. } Finally, cut-distances between the individual distance matrices and a reference matrix offer an additional characterization of their global structure. We compute cut-distances with respect to a zero matrix (the cut-norm), a random and an homogeneous matrix (defined as in the last paragraph), and with respect to shuffled versions of themselves, averaged over 100 iterations (cf. Figure \ref{Humans}G). 
We find narrow distributions suggesting all individuals share a similar global structure.

\end{document}